\documentclass[12pt]{article}

\usepackage{amsmath}
\usepackage{amssymb}
\usepackage{graphicx}
\usepackage{epstopdf}
\usepackage{color}
\usepackage{xspace}
\usepackage{fullpage}
\usepackage{url}
\usepackage{arydshln}

\usepackage[utf8]{inputenc}
\usepackage[T1]{fontenc}

\newcommand{\gbf}[1] {\mbox{\boldmath${#1}$\unboldmath}}
\newcommand{\negr}[1]{{\bf {#1}}}

\renewcommand{\Vec}[1]{\overrightarrow{#1}}

\newcommand{\U}{{B_1}}

\newcommand{\M}{\mathbf{M}}
\newcommand{\dM}{\det(\mathbf{M})}
\renewcommand{\r}{\rho}
\renewcommand{\t}{\alpha}
\renewcommand{\a}{{\beta}}

\newcommand{\R}{\mathbf{L}}
\newcommand{\X}{\mathbf{X}}
\newcommand{\E}{\mathbf{E}}
\newcommand{\J}{\det(\mathbf{J})}

\newcommand{\DV}{DV}

\newcommand{\fgb}{\textsc{FGb}\xspace}

\newcommand{\fgbelim}{\texttt{fgbrs:-fgb\_gbasis}\xspace}

\newcommand{\rs}{\textsc{RS}\xspace}
\newcommand{\rsisolate}{\texttt{RootFinding:-Isolate}\xspace}

\newcommand{\maple}{\textsc{maple}\xspace}
\newcommand{\x}{\ensuremath{x}}

\newtheorem{defi}{Definition}
\newtheorem{rem}{Remark}
\newtheorem{nota}{Notations}
\newtheorem{theo}{Theorem}

\newenvironment{definition}{\begin{defi}}{\end{defi}}
\newenvironment{notations}{\begin{nota}\ }{\end{nota}}
\newenvironment{remark}{\begin{rem}}{\end{rem}}
\newenvironment{theorem}{\begin{theo}\ }{\end{theo}}

\title{On the determination of cusp points of 3-R\underline{P}R parallel manipulators}
\author{%
{G. Moroz$^*$, F. Rouiller$^*$, D. Chablat$^{**}$, P. Wenger$^{**}$}%
\vspace{1.6mm}\\
{\small $^{*}$Laboratoire d'informatique de Paris}, \\
{\small104 avenue du Pr\'esident Kennedy, 75016 Paris, France,} \\
{\small$^{**}$Institut de Recherche en Communications et Cybern\'etique de Nantes,} \\
{\small UMR CNRS 6597, 1 rue de la Noe, 44321 Nantes, France}\\\
{\small Guillaume.Moroz@lip6.fr\quad Fabrice.Rouiller@inria.fr} \\
{\small Damien.Chablat@irccyn.ec-nantes.fr \quad  Philippe.Wenger@irccyn.ec-nantes.fr }
}

\begin{document}
\maketitle
\subsection*{\centering Abstract}
{\em This paper investigates the cuspidal configurations of 3-R\underline{P}R parallel manipulators that may appear on their singular surfaces in the joint space. Cusp points play an important role in the kinematic behavior of parallel manipulators since they make possible a non-singular change of assembly mode. In previous works, the cusp points were calculated in sections of the joint space by solving a 24th-degree polynomial without any proof that this polynomial was the only one that gives all solutions. The purpose of this study is to propose a rigorous methodology to determine the cusp points of 3-R\underline{P}R manipulators and to certify that all cusp points are found. This methodology uses the notion of discriminant varieties and resorts to Gr\"obner bases for the solutions of systems of equations.}
\begin{keyword}
Kinematics, Singularities, Cusp, Parallel manipulator, Symbolic computation
\end{keyword}
\section{Introduction}
Because at a singularity a parallel manipulator loses its stiffness, it is of primary importance to be able to characterize these special configurations. This is, however, a very challenging task for a general parallel manipulator. Planar parallel manipulators have received a lot of attention [1-5] because of their relative simplicity with respect to their spatial counterparts. Moreover, studying the former may help understand the latter. Planar manipulators with three extensible leg-rods, referred to as 3-R\underline{P}R manipulators, have often been studied. Such manipulators may have up to six assembly modes and their direct kinematics can be written in a polynomial of degree six \cite{GMmmt94,Wjmd98}. Moreover, they may have singularities (configurations where two direct kinematic solutions coincide). It was first pointed out that to move from one assembly mode to another, the manipulator should cross a singularity \cite{BLieee86,HPmmt93}. Later, Innocenti and Parenti-Castelli \cite{IPC98} showed, using numerical experiments, that this statement is not true in general. In fact, this statement is only true under some special geometric conditions, such as similar base and mobile platforms \cite{McaD99,KGjfr05}. More recently, Macho et al. \cite{MAPHiftomm07} proposed a method to plan non-singular assembly-mode changing trajectories. McAree \cite{McaD99} pointed out that for a 3-R\underline{P}R parallel manipulator, if a point with triple direct kinematic solutions exists in the joint space, then the nonsingular change of assembly mode is possible. This result holds under some assumptions on the topology of the singularities \cite{Hck09}. For other mechanisms than 3-R\underline{P}R manipulator, it is also interesting to note that encircling a cusp  point is not the only way to execute a non-singular change of assembly mode \cite{BWSieee08}. A condition for three direct kinematic solutions to coincide was established in \cite{McaD99}. This condition was then exploited in \cite{ZWCr07} to derive a univariate polynomial of degree 96. A factored expression  was obtained, one of the factors being a 24th-degree polynomial. The authors observed on many examples that the cusp points were each time defined by the 24th-degree polynomial, the remaining factors always defining spurious solutions only. However, they did not attempt to certify that the 24th-degree polynomial was really the only valid factor. Moreover, they never found more than 8 cusp points.

The purpose of this paper is to propose a rigorous methodology to determine the cusp points of any 3-R\underline{P}R manipulator and to certify that all cusp points are found. This methodology uses the notion of discriminant varieties and resorts to Gr\"obner bases to solve the systems of equations. Using symbolic computation, it is verified that the cusp points are really defined by a 24th-degree polynomial. For any given 3-R\underline{P}R manipulator geometry, the maximum number of cusp points in sectional sections of the joint space is determined and the results are certified. In particular, a robot with 10 cusp points in a cross section of its joint space is found for the first time.

The following section introduces the manipulators studied and recalls the main known results. Section 3 describes the algebraic tools used. Last section presents the methodology that is proposed to determine the cusp points using the algebraic tools.
\section{Modeling and state of the art}
\label{partieI}
\subsection{Robot studied and Modeling}
\begin{figure}[hbt]
  \begin{center}
        \includegraphics[scale=1]{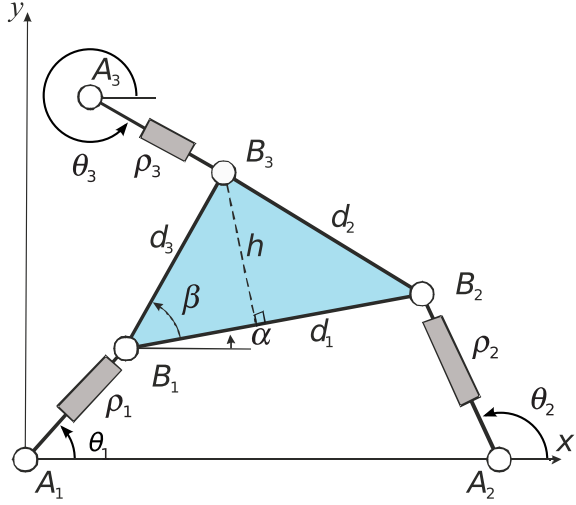}
        \caption{The 3-R\underline{P}R parallel manipulator under study.}
        \protect\label{figure:manipulator_general}
  \end{center}
\end{figure}

A general 3-R\underline{P}R planar parallel manipulator is shown in Figure 1. This manipulator has three extensible leg-rods actuated with prismatic joints. This manipulator can move its moving platform $B_1B_2B_3$ in the plane. The vector of the three leg-rod lengths is $\gbf{L}\equiv (\rho_1, \rho_2, \rho_3)$ (Figure \ref{figure:manipulator_general}). The geometric parameters of the manipulators are the three sides of the moving platform $d_1$, $d_2$, $d_3$ and the position of the base revolute joint centers $A_1$, $A_2$ and $A_3$. The reference frame is centered at $A_1$ and the x-axis passes through $A_2$. Thus, $A_1 = (0, 0)$, $A_2 = (A_{2x}, 0)$ and $A_3 = (A_{3x} , A_{3y})$.

\begin{table}[h]
\begin{center}
    \begin{tabular}{|c|c|c|c|}
    \hline
       & Parameters & Position variables & Equation constraints \\
     \hline
     MD &  $\rho_1, \rho_2, \rho_3$ & $\theta_1, \theta_2, \theta_3$ &
        $\left\{\begin{array}{l}
            \|\Vec{B_1 B_2}\|=d_1\\
            \|\Vec{B_2 B_3}\|=d_2\\
            \|\Vec{B_3 B_1}\|=d_3
        \end{array}\right.$
        \\
     \hline
     GRS &  $\rho_1, \rho_2, \rho_3$ & ${B_1}_x, {B_1}_y, \alpha$ &
        $\left\{\begin{array}{l}
            \|\Vec{A_1 B_1}\|=\r_1\\
            \|\Vec{A_2 B_2}\|=\r_2\\
            \|\Vec{A_3 B_3}\|=\r_3
        \end{array}\right.$
        \\
     \hline
    \end{tabular}
\end{center}
\caption{Two models for the 3-R\underline{P}R manipulator}
\label{models}
\end{table}

The position of the moving platform can be expressed by two different sets of variables (see also Table \ref{models}):
\begin{itemize}
\item \textbf{MD Model} (McAree-Daniel): the 3 angles $\theta_1, \theta_2, \theta_3$.
\item \textbf{GSR Model} (Gosselin-Sefriou-Richard): the position coordinates of point $B_1$ (${B_1}_x, {B_1}_y$) and the platform orientation $\alpha$, ($\alpha_x = cos(\alpha), \alpha_y = sin(\alpha)$).
\end{itemize}

Each of these sets of variables leads to a different modeling system.
\subsubsection{MD Model}
The first model
was used by MacAree and Daniel in \cite{McaD99} and has been extensively studied in \cite{ZWCr07}.
Let $\gbf{\theta} \equiv (\theta_1, \theta_2, \theta_3)$ define the three angles between the leg-rods and the x-axis. The six parameters (\gbf{L}, \gbf{\theta}) define a configuration of the manipulator but only three of them are independent, so that the configuration space is a 3-dimensional manifold embedded in a 6-dimensional space. The dependency between (\gbf{L}, \gbf{\theta}) is obtained by writing the constraint equations of the manipulator, namely, the fixed distances between the three vertices of the mobile platform $B_1, B_2, B_3$:
\begin{equation}
\gbf{\Gamma}
\begin{cases}
  \Gamma_1(\gbf{L}, \gbf{\theta})= [\negr b_2 (\gbf{L}, \gbf{\theta})- \negr b_1 (\gbf{L}, \gbf{\theta})]^T
                                   [\negr b_2 (\gbf{L}, \gbf{\theta})- \negr b_1 (\gbf{L}, \gbf{\theta})] -d_1^2\\
  \Gamma_2(\gbf{L}, \gbf{\theta})= [\negr b_3 (\gbf{L}, \gbf{\theta})- \negr b_2 (\gbf{L}, \gbf{\theta})]^T
                                   [\negr b_3 (\gbf{L}, \gbf{\theta})- \negr b_2 (\gbf{L}, \gbf{\theta})] -d_2^2\\
  \Gamma_3(\gbf{L}, \gbf{\theta})= [\negr b_1 (\gbf{L}, \gbf{\theta})- \negr b_3 (\gbf{L}, \gbf{\theta})]^T
                                   [\negr b_1 (\gbf{L}, \gbf{\theta})- \negr b_3 (\gbf{L}, \gbf{\theta})] -d_3^2\\
  \end{cases}
  \label{MDequ}
\end{equation}
where $\negr b_i$ is the vector defining the coordinates of $B_i$ in the reference frame as function of $\negr L$ and \gbf{\theta}. 
\subsubsection{GSR Model}
\label{GSR}
The GRS model was introduced in \cite{GRSmmt92}. This model, combined with appropriate algebraic
methods (see Section \ref{algtools}), will allow us to use a full symbolic computation of all cuspidal
configurations. In this modeling, point $B_1$ and angle $\alpha$ are used to specify the pose and the orientation of the platform. Let $\X \equiv ( {B_1}_x, {B_1}_y, \alpha_x, \alpha_y )$, where $\alpha_x$ (resp. $\alpha_y$) denotes $cos(\alpha)$ (resp. $sin(\alpha)$). Let $\beta_x$ (resp. $\beta_y$) denote $cos(\beta)$ (resp. $sin(\beta)$). With these variables, we can parametrize the positions of the three points $B_1, B_2$ and $B_3$ as follows:
\begin{align*}
\negr{b_1}&\left|\begin{array}{c} {B_1}_x\\ {B_1}_y\end{array}\right. &
\negr{b_2}&\left|\begin{array}{c} {B_1}_x+d_1\alpha_x\\ {B_1}_y+d_1\alpha_y\end{array}\right. &
\negr{b_3}&\left|\begin{array}{c} {B_1}_x+d_3(\alpha_x\beta_x-\alpha_y\beta_y)\\ {B_1}_y+d_3(\alpha_x\beta_y+\alpha_y\beta_x)\end{array}\right.
\end{align*}

The lengths $\r_1,\r_2,\r_3$ define the geometric constraints of the robot, and can be written as the euclidian norm of the vectors $\Vec{A_1B_1},\Vec{A_2B_2},\Vec{A_3B_3}$. Finally, the dependency between $(\gbf{L},\gbf{\X})$ can be identified by writing the distances between the vertices of the moving platform and the vertices of the base platform. These distances $\r_1,\r_2,\r_3$ are the norms of the vectors $\Vec{A_1B_1},\Vec{A_2B_2},\Vec{A_3B_3}$:

\begin{align}
\begin{cases}
    E_1 (\gbf{L}, \gbf{X}) =
        [\negr b_1 (\gbf{L}, \gbf{X})- \negr a_1 ]^T
        [\negr b_1 (\gbf{L}, \gbf{X})- \negr a_1 ] -\r_1^2\\
    E_2 (\gbf{L}, \gbf{X}) =
        [\negr b_2 (\gbf{L}, \gbf{X})- \negr a_2 ]^T
        [\negr b_2 (\gbf{L}, \gbf{X})- \negr a_2 ] -\r_2^2\\
    E_3 (\gbf{L}, \gbf{X}) =
        [\negr b_3 (\gbf{L}, \gbf{X})- \negr a_3 ]^T
        [\negr b_3 (\gbf{L}, \gbf{X})- \negr a_3 ] -\r_3^2\\
\end{cases}
  \label{static}
\end{align}
where $\negr a_i$ is the vector defining the constant coordinates of $A_i$ in the reference frame.


The formulas in Equations \ref{static} can be expanded as follows:
\begin{equation}
\label{constraints}
\E
\left\{
\begin{array}{lcl}
    {B_1}_x^2+ {B_1}_y^2-\r_1^2&=&0\\
    ({B_1}_x+d_1\t_x-{A_2}_x)^2+({B_1}_y+d_1\t_y)^2-\r_2^2&=&0\\
    ({B_1}_x+d_3\t_x\a_x-d_3\t_y\a_y-{A_3}_x)^2+({B_1}_y+d_3\t_x\a_y+d_3\t_y\a_x-{A_3}_y)^2-\r_3^2&=&0\\
    \t_x^2+\t_y^2-1&=&0\\
\end{array}
\right.
\end{equation}

The last polynomial in $\E$ comes from the variable $\alpha$ replaced by the $2$ variables $\t_x,\t_y$ and the equation constraint $\t_x^2+\t_y^2-1=0$. This change of variables allows us to avoid the sine and cosine functions and keep the system algebraic. Moreover, this change of variables does not introduce any spurious solutions since the function
$$\begin{array}{ll}[-\pi, \pi[ & \rightarrow \mathcal{C}\\ \theta & \mapsto (\cos(\theta),\sin(\theta)) \end{array}$$
is bijective on the circle $\mathcal{C}$.

The following subsections recall the main properties of the 3-R\underline{P}R manipulator observed using the MD Model.

\subsection{Properties observed}
\label{coalescence}
In a special configuration (\gbf{L}, \gbf{\theta}), the manipulator meets a singularity, where two assembly modes coalesce. This happens when the constraint Jacobian  drops rank. A 3-R\underline{P}R parallel manipulator is in a singularity whenever the three lines $(A_iB_i), i=1..3$, are concurrent or parallel \cite{BZGjdm03}.

When three assembly modes coalesce, the manipulator is said to be in a cuspidal con\-fi\-gu\-ra\-tion. Using series expansion, McAree and Daniel show that in such configurations, the manipulator loses first and second order constraints. This gives a necessary condition for a manipulator to be in a cuspidal configuration (see Table \ref{algorithms}).

\begin{figure}
   \begin{center}
   \includegraphics[scale=1]{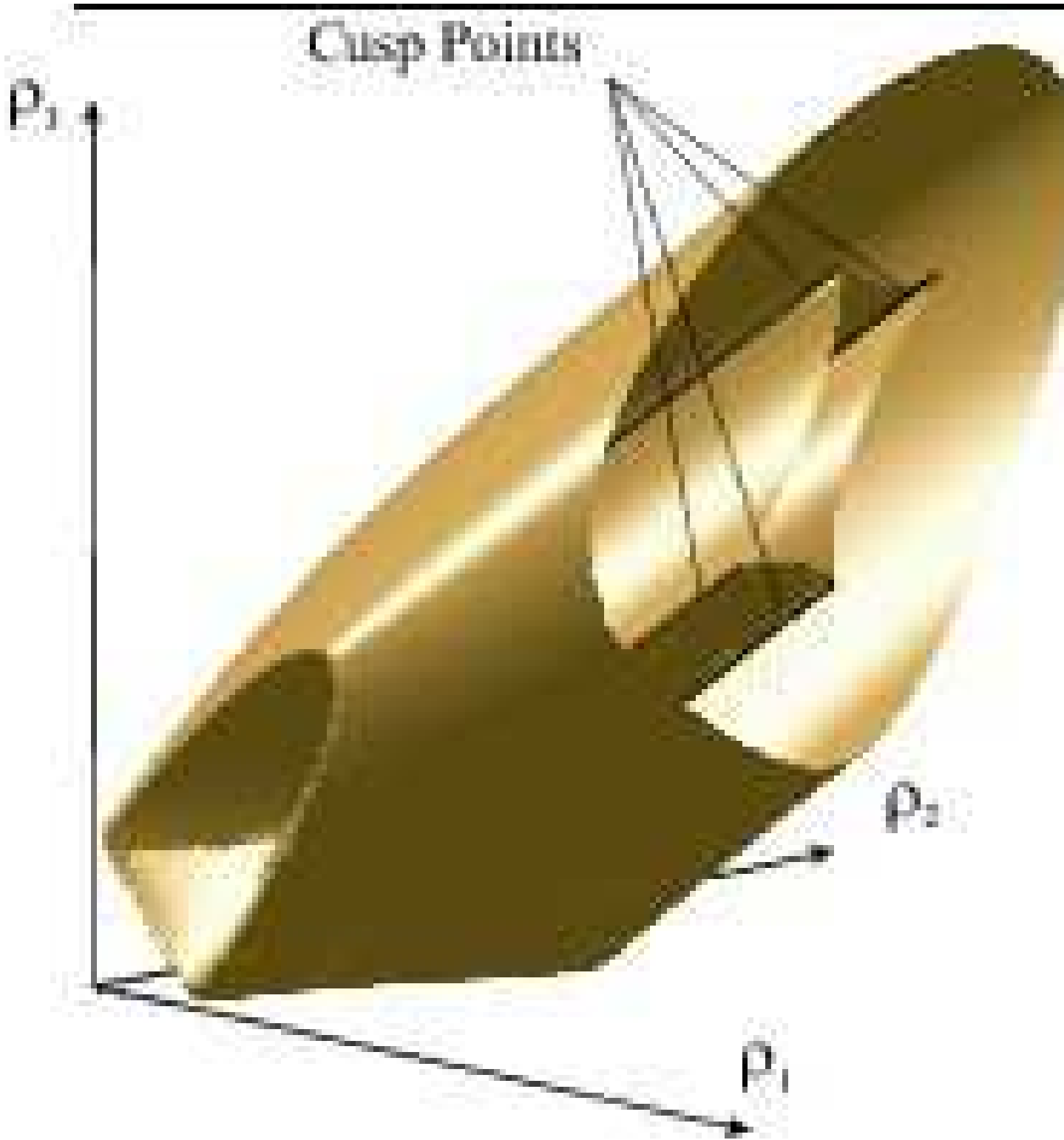}
   \caption{Joint space singularity surfaces of the 3-R\underline{P}R manipulator studied when $\rho_1$ varies from 0 to 5 \cite{ZWCr07}.}
   \label{fig:surface_singuliere}
   \end{center}
\end{figure}

Based on series expansion, \cite{ZWCr07}  provided an algorithm to plot the singular curves in sectional slices of the joint space and to calculate the cusp points that appear on these curves. They calculated the cusp points in various sectional slices of the joint space for the manipulator defined by the following geometric parameters:
\begin{equation}
  A_1=(0, 0) \quad A_2=(15.91, 0) \quad A_3=(0, 10) \quad d_1=17.04 \quad d_2=16.54 \quad d_3=20.84
  \label{geometrical}
\end{equation}
The following facts were observed in \cite{ZWCr07}: 
\begin{itemize}
\item no more than 8 cusp points were found in a sectional slice defined for a given value of $\rho_1$; this result was also observed in many other manipulator examples;
\item the number of cusp points per section stabilizes to four when $\rho_1$ exceeds a certain value;
\item the set of all cusps points defined a set of curves in the 3-dimensional joint space (Fig.~\ref{fig:surface_singuliere}).
\end{itemize}
\subsection{Limitations of the previous study}
\begin{figure}
  \begin{center}
  \parbox[t]{0.49\columnwidth}{\centering (a)
  \begin{tabular}[t]{l}
  $A2x=11$, $A3x=7$, $A3y=10$\\
  $d_1=5$, $d_3=5$, $\beta=37$\\
  $\theta_1=53$, $\theta_2=127$, $\theta_3=-90$\\
  $\r_1=5$, $\r_2=5$, $\r_3=3$
  \end{tabular}\\
  \includegraphics[width=0.4\textwidth]{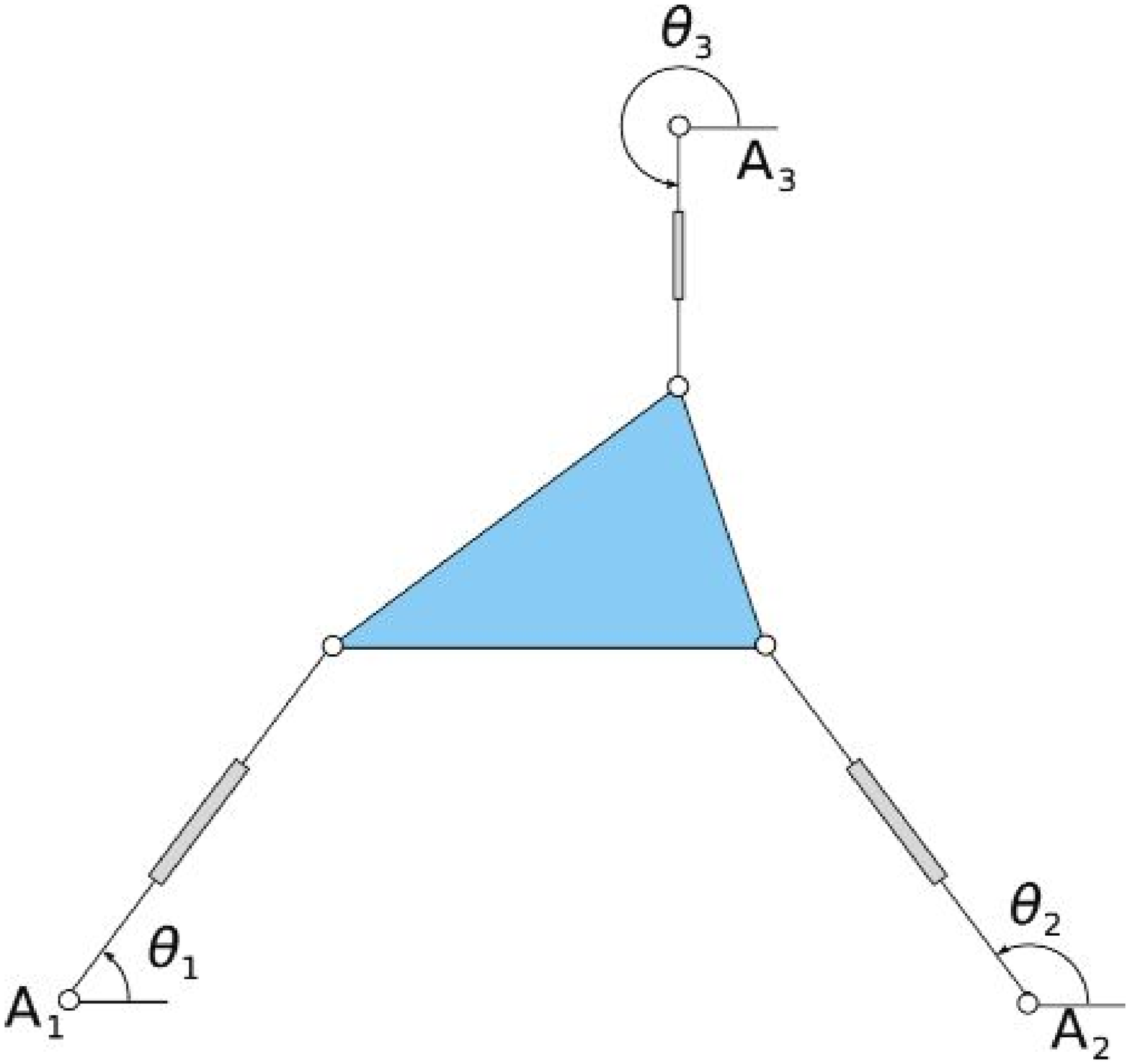}}
  \parbox[t]{0.49\columnwidth}{\centering (b)
  \begin{tabular}[t]{l}
  $A2x=11$, $A3x=7$, $A3y=10$\\
  $d_1=5$, $d_3=5$, $\beta=-37$\\
  $\theta_1=53$, $\theta_2=127$, $\theta_3=-90$\\
  $\r_1=5$, $\r_2=5$, $\r_3=9$
  \end{tabular}\\
  \includegraphics[width=0.4\textwidth]{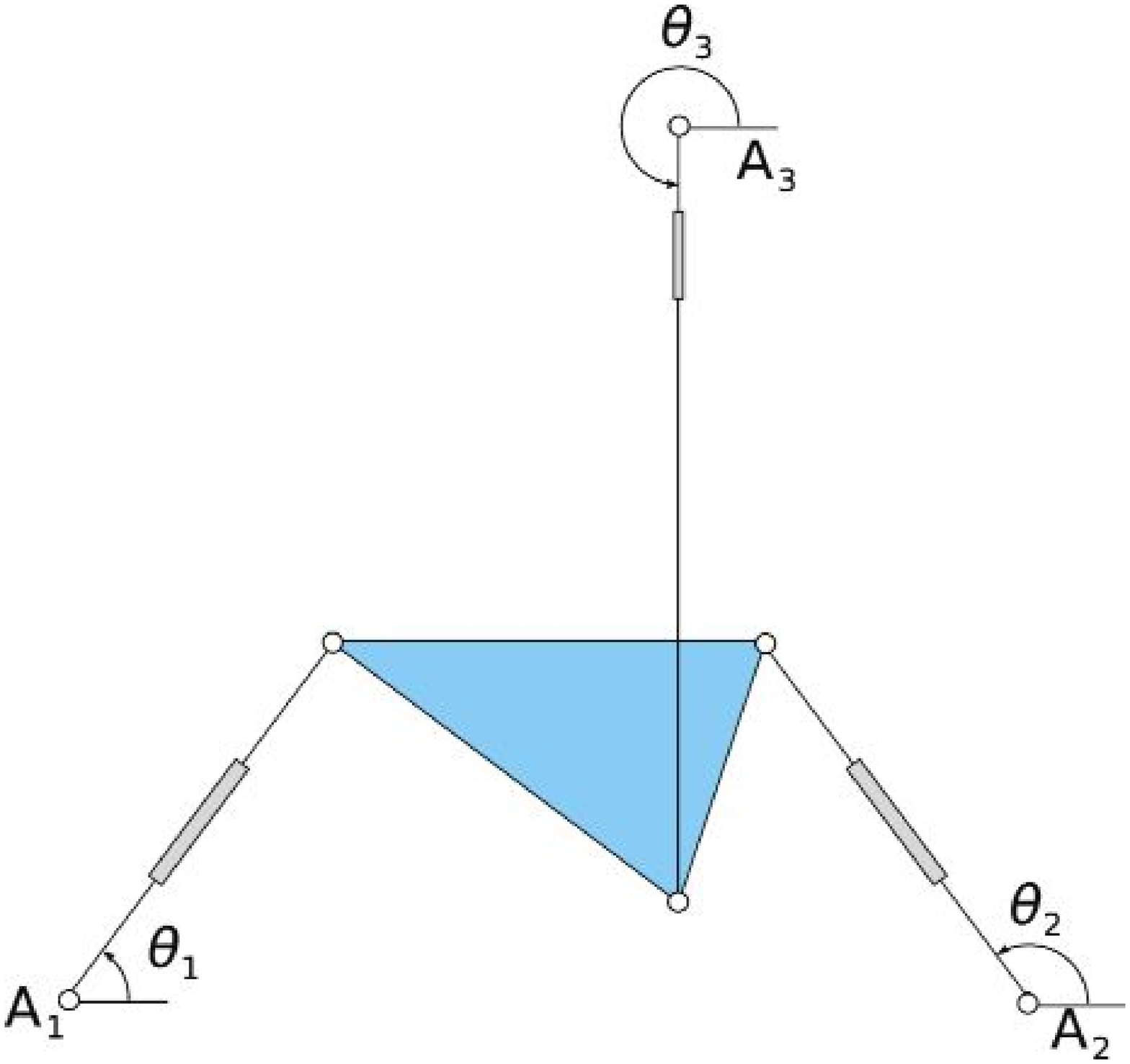}}
\caption{A platform triangle (left) and its mirror image (right)}
\label{triangles}
\end{center}
\end{figure}

In \cite{ZWCr07},  the cusp points were calculated in any sectional slice of the joint space by resorting to complex algebraic calculus based on expansion and elimination procedures. A univariate polynomial of degree 96 was obtained, which was shown to factor into several polynomials, one of which is of degree 24 and was conjectured to be the one that gives all cusp points in the sectional slice at hand. The other factors were found to give always spurious solutions.
The following remarks can be drawn:
\begin{itemize}
\item there is no proof that the above mentioned 24-degree polynomial is indeed the only polynomial that gives all solutions;
\item the sectional slices were analyzed by a discrete method, and, thus, the analysis was not exhaustive;
\item a maximum number of 8 cusps points were found in many manipulator examples, but there is no proof that no more than 8 solutions can be obtained.
\item   previous equation modeling does not distinguish between a platform triangle and its mirror image (see Figure \ref{triangles}), i.e. a triangle with the same $d_i$ but with an angle $\beta$ of opposite sign.
\end{itemize}

One limitation of the previous studies came from the algebraic tools used to describe the robot, which introduced spurious components that had to be removed with empirical observations. In the literature, another method exists based on the elimination of variables, reducing the problem to the computation of the triple roots of an univariate polynomial \cite{OW95,CR02,Wjmd98}. Unfortunately, due to the size of the univariate polynomial, this method ran out of memory when applied to 3-R\underline{P}R manipulators \cite{ZWCr07}.

The following section presents the minimal Discriminant Variety and the Jacobian criterion. These algebraic tools allowed us to compute and describe exactly the cuspidal configurations of the 3-R\underline{P}R manipulators, without introducing any spurious components.

\section{Algebraic tools}
\label{algtools}
This section recalls some mathematic definitions and their implementation in Maple software. The parametric systems we consider are supposed to have a finite number of complex solutions for almost all parameter values. This property is checked by computing a Gröbner basis of the system at hand (see \cite[page 274]{BWbook93} and \cite[Theorem 2]{LR07} for more details).
\subsection{Discriminant Variety}
A Discriminant Variety $V$ \cite{LR07} is associated to a parametric polynomial system $S$. One of its main properties is that the real roots of $S$ can be parametrized continuously on each open connected set in the complement of $V$.


    \pagebreak

\begin{notations}

    In this section, we denote by $S(t_1,...,t_s)$ a polynomial system of the form:
        $$p_1=0,...,p_m=0 \quad and\quad q_1>0,...,q_k>0$$
depending on the $s$ parameters $t_1,...t_s$ and the $n$ variables $x_1,...,x_n$.

    For more simplicity, $(t_1,...,t_s)$ will be omitted in the subsequent sections.
\end{notations}

For example, if we want to describe the number of solutions to the direct kinematics problem according to the articular parameters, then:
\begin{itemize}
\item the polynomial equations $p_1,...,p_m$ are equations (\ref{constraints}) with the specifications (\ref{geometrical})
\item the polynomial inequations $q_1>0,...,q_k>0$ are $\r_1>0, \r_2>0, \r_3>0$
\item the parameters $t_i$ are the articular variables $\r_1,\r_2,\r_3$ and the variables $x_i$ are the pose variables ${B_1}_x,{B_1}_y,\t_x,\t_y$.
\end{itemize}

The following definition is adapted from \cite{LR07} for our problem.

\begin{definition}(Discriminant Variety)\\
  Let $S$ be any parametric system. With the notation defined above, we call \emph{Discriminant Variety} of $S$ a set $V$ of parameter values such that on each connected part of $\mathbb{R}^s\setminus V$:
\begin{itemize}
\item[i.] the roots of $S$ can be parametrized as continuous functions of the parameters;
\item[ii.] the roots of $S$ do not cross.
\end{itemize}

Moreover, the intersection of all the discriminant varieties is called the \emph{minimal} discriminant variety and is denoted by $\DV$.
\end{definition}
\begin{remark}
The conditions i. and ii. imply that for every open set in the complement of $\DV$, the number of solutions is constant.
\end{remark}

One of the main property of the minimal discriminant variety $\DV$ is that if $S$ has finitely solutions for almost all parameter values, then for each connected cell $\mathcal{U}$ of $\mathbb{R}^s\setminus\DV$, the number of solutions of $S$ restricted to $\mathcal{U}$ is constant.

In the problem at hand, the minimal discriminant variety may be split into two components:
\begin{itemize}
  \item[-] the inequation bounds in the parameters space. As far as a 3-R\underline{P}R manipulator is considered, these bounds are trivial since the inequations already depend solely on the parameters;
  \item[-] the projection of the parallel singularities (defined in \cite{GAieee90}, see also section \ref{jacobiancriterion}). These values are computed as follows:
    \begin{itemize}
      \item[1.] Compute $\M_1,\ldots,\M_{\binom{k}{n}}$ as the $n\times n$ minors of the Jacobian matrix:
$$\left( \begin{array}{ccc}
  \dfrac{\partial p_1}{\partial x_1}&\cdots &\dfrac{\partial p_1}{\partial x_n}\\
  \vdots& & \vdots\\
  \vdots& & \vdots\\
  \dfrac{\partial p_k}{\partial x_1}&\cdots &\dfrac{\partial p_k}{\partial x_n}\\
\end{array}\right)
  $$
\item[2.] Eliminate the variables $x_1,\ldots,x_n$ in the system
    $$p_1=0,\ldots,p_k=0,\dM_1=0,\ldots,\dM_{\binom{k}{n}}=0$$
    The zero iso-surface of the returned polynomials is the projection of the desired singularities. 
    \end{itemize}
\end{itemize}
\subsection{Degree of a $0$-dimensional system and multiplicity of a root}
\label{degree}
The interested reader may find more detailed definitions on the \emph{degree} and \emph{multiplicity} in \cite{M06} or in \cite[chapter 4]{CLO98}.

\begin{definition}(degree)
  Given a system $S$ of polynomial equations with finitely many complex solutions,  the degree of $S$ is exactly the number of roots counted with multiplicities.
\end{definition}

The following unusual definition is exactly the same as the classical one \cite{CLO98} (dimension of a local ring), the reader will find the correspondence in \cite{Raaecc99}.

\begin{definition}(multiplicity)
  Let $S$ be a system of polynomial equations in $\x_1,...,\x_n$ with finitely many complex solutions. Let $l:\mathbb{C}^n\rightarrow\mathbb{C}$ be a linear form and $P(Z)$ a univariate polynomial returned by the computation of a rational univariate representation \cite{Raaecc99}.

Then $P$ can be written as:
  $$ P(Z) = \prod_{\alpha\ root\ of\ S} (Z-l(\alpha))^{\mu_\alpha}$$
 
  We call \emph{multiplicity} of $\alpha$ the exponent $\mu_\alpha$.
\end{definition}
\begin{remark}
  When $S$ is reduced to a univariate polynomial equation in X with $k$ roots
  \begin{equation*}
    \prod_{i=1}^k(X-c_i)^{\mu_i} =0
\end{equation*}
then the multiplicity of each root $c_i$ is exactly $\mu_i$.
\end{remark}

By extension, we can also define the multiplicity of the roots of a parametric system.

\begin{definition}(multiplicity in a parametric system)
  Let $S(t_1,...,t_s)$ be a parametric system depending on the parameters $t_1,...,t_s$ and on the variables $\x_1...,\x_n$. Assume furthermore that for all parameter values $(t_1^0,...,t_s^0)\in\mathbb{C}^s$, the number of complex roots of $S(t_1^0,...,t_s^0)$ is finite.

Then, the \emph{multiplicity} of a root $\mathbf{\alpha}=(t_1^0,...,t_s^0,\x_1^0,...,\x_n^0)$ is the multiplicity of the root $(\x_1^0,...,\x_n^0)$ in the system $S(t_1^0,...,t_s^0)$.

\end{definition}
\begin{remark}
This multiplicity definition can be naturally extended to define the multiplicity of a leaf above a connected component in the complement of the discriminant variety.
\end{remark}

\subsection{Jacobian criterion}
\label{jacobiancriterion}

We recall here the Jacobian criterion, used in this article to extract the points of multiplicity greater than or equal to $2$. More details can be found in \cite{E94}. The Jacobian matrix to be considered here is exactly the same as the one used in \cite{GAieee90} to define the singularities of type 2 of a manipulator. In this article, we will refer to these type-2 singularities as \emph{parallel singularities}.

\begin{theorem}
  Let $p_1,\ldots,p_m$ be parametric polynomials of $\mathbb{C}[t_1,...,t_s][\x_1,\ldots,\x_n]$ admitting a finite set of common complex roots for each value of the parameters $t_1,...,t_s$. Let $J$ be the Jacobian matrix related to the variables $\x_1,...,\x_n$:
  $$\left[\begin{array}{ccc}
    \dfrac{\partial p_1}{\partial \x_1} & \cdots & \dfrac{\partial p_1}{\partial \x_n}\\
    \vdots & & \vdots\\
    \vdots & & \vdots\\
    \dfrac{\partial p_m}{\partial \x_1} & \cdots & \dfrac{\partial p_m}{\partial \x_n}\\
  \end{array}
  \right]$$
  If $\M_1,\ldots,\M_{\binom{m}{n}}$ denotes the $n\times n$ minors of $J$, then for all $\mathbf{t^0}=(t_1^0,...,t_s^0)\in\mathbb{C}^s$, the following parametric system:
  $$\begin{cases}
   p_1(\mathbf{t^0},\x_1,...,\x_n)=0,\ldots,p_m(\mathbf{t^0},\x_1,...,\x_n)=0\\
   \dM_1(\mathbf{t^0},\x_1,...,\x_n)=0,\ldots,\dM_{\binom{m}{n}}(\mathbf{t^0},\x_1,...,\x_n)=0
  \end{cases}$$
  has a solution in $\x_1,...,\x_n$ if and only if the system
  $$p_1(\mathbf{t^0},\x_1,...,\x_n)=0,\ldots,p_m(\mathbf{t^0},\x_1,...,\x_n)=0$$
  has at least one root of multiplicity greater than or equal to $2$.
\end{theorem}

This theorem can be used to define singular configurations as in \cite{GAieee90}, but also to define the cuspidal configurations, as shown in section \ref{Cuspidal}.

\subsection{Software implementation in Maple}
\label{soft}
The algebraic tools used in this paper were adapted from the SALSA library developed at INRIA. The manipulations and computations of the algebraic objects are performed with the math software \maple. The functions used are described below.

\paragraph{\fgb}
Given a set of polynomials generating an ideal $I$, \fgbelim allows the computation of a system of generators of the ideal. These generators provide a way to reduce any polynomial to a canonical unique form modulo $I$.
With the option \texttt{``elim''=true}, it makes it possible  to compute an algebraic system, the zero of which is the closure of the projection of the roots of $I$.

This elimination of variables is based on Gr\"obner basis computations, using the algorithms F4 (\cite{Fjpaa99}). As explained in the following sections, this function is used to compute the singular and cuspidal positions, as well as the discriminant variety.
\paragraph{\rs}
A system $S$ of polynomial equations $p_1=0,\ldots,p_k=0$ is said $0$-dimensional if its number of complex solutions is finite (this can be straightforwardly tested on a Gr\"obner basis of the system).

In this case, the \rsisolate function computes directly the real solutions of $S$. By default, each solution is given by a box with rational bounds, ensuring that no pairs of boxes overlaps. The algorithms implemented in Maple are based on the Rational Univariate Representation (\cite{Raaecc99}).

This function allows us to compute the number of cuspidal positions.

\paragraph{Discriminant Variety}
Given a parametric system of equations $p_1=0,\ldots,p_k=0$ and inequations $g_1\neq 0,\ldots,g_r\neq 0$, $t_1,\ldots,t_s$ being the parameters and $x_1,\ldots,x_n$ the unknowns, its minimal discriminant variety is computed with the algorithm presented in \cite{LR07}.

As shown further in section \ref{DV}, the discriminant variety is a key tool for the description of the cuspidal positions.

This function is part of the package \texttt{RootFinding[Parametric]} in \maple.

\paragraph{Limitations}
These functions stand only for polynomials with rational coefficients. In particular, the trigonometric functions $sin, cos, tan$ have to be replaced by algebraic variables.


\section{New modelling and method}

In this section and the subsequent ones, we use the equation system and the notations induced by the GSR Model presented in section \ref{GSR}. In this model, the position of the platform is given by $4$ variables: ${B_1}_x, {B_1}_y$ the coordinates of the point $B_1$ and $\t_x, \t_y$, the cosine and sine of the angle $\alpha$ respectively.

The configurations of a 3-R\underline{P}R manipulator may be classified into three categories: the regular configurations, the singular configurations and the cuspidal configurations. The description of the special configurations (singular and cuspidal) is important for path planning. First, the regular configurations are the most common ones: these are the configurations where the position of the platform is locally uniquely determined by the values of the control parameters. Then the singular (resp. cuspidal) configurations are two (resp. three) coalesced configurations. These configurations were studied through local analysis in the past, but they can be also described with algebraic methods, as shown further in this article.
\subsection{Singular configurations}
\label{Singular}

From a geometrical point of view, we saw in Section \ref{coalescence} that a singular configuration represents  coalesced assembly modes.

From an algebraic point of view, this condition can be retrieved in term of multiplicities. When the variables $\R$ is specialized with real numbers, the set of equations (\ref{constraints}) has finitely many solutions. Moreover, a multiplicity is associated with each solution (see Section \ref{degree}). Using this notion, it can be verified that the singular assembly modes are exactly configurations that are solutions of multiplicity greater than or equal to $2$. Indeed, the Jacobian criterion \cite{E94} states directly that these configurations are parallel singularities.\\

\subsubsection{Modelling}
To describe the singular points, we use the notion of multiplicity in a $0$-dimensional system. The system of equations (\ref{constraints}) is $0$-dimensional when the lengths $\r_1,\r_2,\r_3$ are given: it has a finite number of complex solutions. In this case, each solution can be associated with an integer greater than or equal to $1$: its multiplicity, as defined in section \ref{degree}.

\begin{definition}
  A configuration $P_0=(\r_1^0, \r_2^0, \r_3^0, {B_1}_x^0,{B_1}_y^0, \t_x^0, \t_y^0)$ of the 3-R\underline{P}R manipulator is said singular if and only if $({B_1}_x^0,{B_1}_y^0,\t_x^0,\t_y^0)$ is a solution of multiplicity greater than or equal to $2$ of:
$$
  \E((\r_1^0,\r_2^0,\r_3^0),\X)=\left[\begin{array}{cccc}0&0&0&0\end{array}\right]^T
$$
\end{definition}

To find the singular configurations, we use the Jacobian criterion, reminded in section \ref{algtools}.

Applying the Jacobian criterion \cite{E94} on the polynomial system Equations (\ref{constraints}), we deduce that the singular configurations are exactly the solutions of the system:
\begin{align}
\begin{cases}
\E(\R,\X)=\left[\begin{array}{cccc}0&0&0&0\end{array}\right]^T\\
\det(\dfrac{\partial\E}{\partial\X})=0
\end{cases}
\label{fullsingular}
\end{align}
\begin{remark}
  \label{rem}
  By adding the determinant of the Jacobian matrix to Equations (\ref{constraints}), we get a new system whose solutions are included in the original system. More precisely, the roots of Equations (\ref{fullsingular}) are exactly the roots of multiplicity greater or equal to $2$ of Equations (\ref{constraints}).
  The solutions to Equations (\ref{constraints}) of multiplicity $2$ are solutions of multiplicity $1$ to Equations (\ref{fullsingular}), and the solutions of multiplicity greater or equal to $3$ in Equations (\ref{constraints}) have a multiplicity greater or equal to $2$ in Equations (\ref{fullsingular}).
\end{remark}

Moreover, using Gr\"obner basis computations to eliminate variables with Maple
, a polynomial in the parametric variables $\R$, denoted $P_{sing}$, can be computed. Its solutions are exactly the sets of length values for which the manipulator admits singular configurations.

\subsubsection{Example}
In this section, the computation results are given for the manipulator example presented in \cite{IPC98,McaD99,ZWCr07,M00}.

\begin{figure}
  \begin{center}
    \includegraphics[width=\columnwidth]{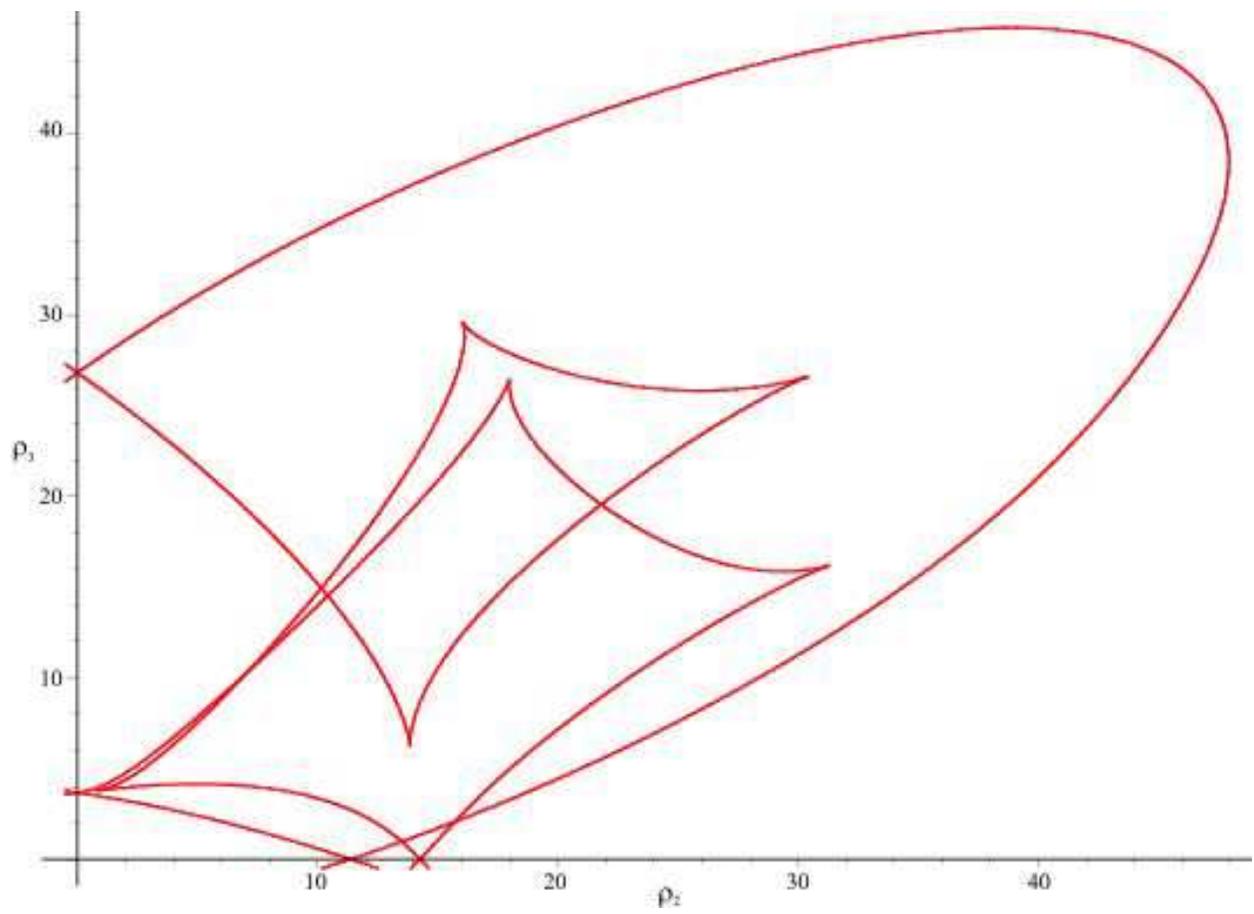}
    \caption{Singular curve for $\r_1=14.98$}
    \label{1498}
  \end{center}
\end{figure}

Once substituting the numerical values of Equations (\ref{geometrical}) into  Equations (\ref{fullsingular}), a system is obtained, the solutions of which are exactly the singular configurations of the studied manipulator. After eliminating the variables $\X$ with Gr\"obner bases computation, a polynomial $P_{sing}$ in $\R$ is obtained of total degree 24. The zeros of this polynomial define a surface in the parameter space. By specifying $\r_1=14.98$ as in \cite{ZWCr07,McaD99}, the curve as in \cite{ZWCr07,McaD99} can be observed in Figure \ref{1498}.

\subsection{Cuspidal configurations}
\label{Cuspidal}
Cuspidal configurations are associated with second-order degeneracies that appear for triply coalesced configurations. As shown in \cite{McaD99,ZWCr07}, these configurations play an important role in path planning.


To find cuspidal configurations, the idea of \cite{McaD99,ZWCr07} was to analyze the kernels of the matrices in the first and second order terms of the series expansion of Equations (\ref{MDequ}). 
It allowed the authors to find  the cuspidal configurations automatically when $\r _1$ is specified as a numerical value. However the description they get is a list of slices of the cuspidal configurations, and they miss what could happen between 2 successive slices. Using the notion of discriminant variety and a generalization of the Jacobian criterion, we introduce a complete certified description of the cuspidal configurations. 
In particular, our approach allows us to find cuspidal configurations that were missed in the previous papers, and to certify that all cuspidal configuration are determined.


\subsubsection{Modelling}
Like for the singular points, we define the cuspidal configurations algebraically.

\begin{definition}
The configuration $P_0=(\r_1^0,\r_2^0,\r_3^0,\U_x^0,\U_y^0,\t_x^0,\t_y^0)$ of the  3-R\underline{P}R manipulator is said cuspidal if and only if $(\U_x^0,\U_y^0,\t_x^0,\t_y^0)$ is a solution of multiplicity $3$ to
$$
\E((\r_1^0,\r_2^0,\r_3^0),\X)=\left[\begin{array}{cccc}0&0&0&0\end{array}\right]^T
$$
.
\end{definition}

In Section \ref{Singular}, the Jacobian criterion allowed us to select the configurations of multiplicity higher than or equal to $2$. However, we now want to compute the configurations of multiplicity $3$. Fortunately, we saw in Remark \ref{rem} that these configurations were the roots of multiplicity $2$ of Equations (\ref{fullsingular}). We can thus use the Jacobian criterion on Equations (\ref{fullsingular}) to get a system whose roots of multiplicity $1$ describe exactly the cuspidal configurations.

Let $\J=\det(\dfrac{\partial\E}{\partial\X})$ and let $\M_1, \M_2, \M_3, \M_4, \M_5$ be the $4\times 4$ minors of the Jacobian matrix of the $5$ polynomials $\E_1,\E_2,\E_3,\E_4, \J$ with respect to the variables $\X$. Then the following system of $9$ equations defines exactly the cuspidal configurations of the 3-R\underline{P}R manipulator:

\begin{align}
\begin{cases}
  \E(\R,\X)=[\begin{array}{cccc}0&0&0&0\end{array}]^T\\
  \J=\dM_1=0\\
    \dM_2=\det\left(\left[
    \begin{array}{cccc}
      \dfrac{\partial\E_1}{\partial\X}&
      \dfrac{\partial\E_2}{\partial\X}&
      \dfrac{\partial\E_3}{\partial\X}&
      \dfrac{\partial\J}{\partial\X}
    \end{array}
    \right]\right)=0\\\\
    \dM_3=\det\left(\left[
    \begin{array}{cccc}
      \dfrac{\partial\E_1}{\partial\X}&
      \dfrac{\partial\E_2}{\partial\X}&
      \dfrac{\partial\E_4}{\partial\X}&
      \dfrac{\partial\J}{\partial\X}
    \end{array}
    \right]\right)=0\\\\
    \dM_4=\det\left(\left[
    \begin{array}{cccc}
      \dfrac{\partial\E_1}{\partial\X}&
      \dfrac{\partial\E_3}{\partial\X}&
      \dfrac{\partial\E_4}{\partial\X}&
      \dfrac{\partial\J}{\partial\X}
    \end{array}
    \right]\right)=0\\\\
    \dM_5=\det\left(\left[
    \begin{array}{cccc}
      \dfrac{\partial\E_2}{\partial\X}&
      \dfrac{\partial\E_3}{\partial\X}&
      \dfrac{\partial\E_4}{\partial\X}&
      \dfrac{\partial\J}{\partial\X}
    \end{array}
    \right]\right)=0
\end{cases}
\label{fullcuspidal}
\end{align}

\begin{remark}
  These equations are algebraically dependent. In particular, even if the number of equations is $9$ and the number of variables is $7$, the set of solutions is a curve of dimension $1$ in $\mathbb{R}^7$.
\end{remark}

System (\ref{fullcuspidal}) contains equations of degree smaller than $5$. This will allow us to compute a certified description of the cuspidal curves in Section \ref{DV}.
\subsubsection{Example}
For the robot example defined by Equations (\ref{geometrical}), triple points are exactly the solutions of Equations (\ref{fullcuspidal}). Moreover, when $\r_1$ is specified, Equations (\ref{fullcuspidal}) has finitely many solutions. Using the methods of real solving for $0$-dimensional systems of the function \texttt{Isolate} of Maple, one can find easily the roots of Equations (\ref{fullcuspidal}) for any given $\r_1$.

The system defined by  Equations (\ref{fullcuspidal}) has 9 equations:
\begin{itemize}
\item the first four equations are the geometric model equations, each being of degree 2;
\item the 5th equation is the Jacobian of these equations and has degree 3;
\item the last four equations are obtained through the iteration of the Jacobian computation, each one is of degree 5..
\end{itemize}

The full equations can be found at the address \url{http://www.irccyn.ec-nantes.fr/~chablat/3RPR_cuspidal.html}.

In particular, for $\r_1=14.98$, we get $6$ triple points (see Table~\ref{cusppoints}), which confirms the results of \cite{ZWCr07}.

\begin{table}
  \begin{center}
\begin{tabular}{|c|c|c|c|c|c|c|}
  \cline{2-7}
  \multicolumn{1}{c|}{}&$\r_2$&$\r_3$&$\U_x$&$\U_y$&$\t_x$&$\t_y$\\
  \hline
1&0.845& 3.777& 5.336& -13.997& 0.633& 0.773\\
  \hline
2& 13.851& 6.260& -14.963& 0.698& 0.998& -0.045\\
  \hline
3& 31.276& 16.178& -6.104& 13.679& -0.543& -0.839\\
  \hline
4& 17.988& 26.446& 14.721& -2.769& -0.985& 0.167\\
  \hline
5& 30.449& 26.619& -10.363& 10.816&.537& 0.843\\
  \hline
6& 16.027& 29.566& 14.437& 3.995& 0.999& -0.010\\
  \hline
\end{tabular}
\caption{The 6 cuspidal configurations (roots of Equations \ref{fullcuspidal}) for $\r_1=14.98$}
\label{cusppoints}
\end{center}
\end{table}

Moreover, in \cite{ZWCr07}, the authors have shown that for different numerical values of $\r_1$, among different spurious factors, it is possible to extract a univariate polynomial in $t_1 := tan(\theta_1)$ of degree24. 

Using our cusp modeling, if we add the new variable $t_1 = \tan(\theta_1) = \U_y/\U_x$ and eliminate the variables $\r_2,\r_3,\U_x,\U_y\t_x,\t_y$ from Equations \ref{fullcuspidal} with Gröbner basis computation, we can compute (in 9 seconds) a bivariate polynomial $Q$ in $\r_1$ and $t_1$. The real roots defined by $Q$ is the closure of the projection of the cuspidal configurations. We can see that $Q$ is irreducible and has a degree $24$ in $t_1$. In particular, by substituting a numerical value of $\r_1$ in $Q$, we get directly a univariate polynomial of degree $24$ in $t_1$. This generalizes and certifies the results in \cite{ZWCr07}.

\subsection{Cusps analysis}
\label{DV}
Equations (\ref{fullcuspidal}) define implicitely the cuspidal configurations of the robot. Without further computations, these equations are not sufficient to describe the geometry of the triple roots of the 3-R\underline{P}R manipulator. In \cite{McaD99,ZWCr07}, the authors observed that the set of cuspidal configurations is finite in slices of the robot joint space for given values of $\r_1$. This leads to the conjecture that the set of cuspidal configurations forms a curve in the full joint space  ($\r_1,\r_2,\r_3$). Furthermore, the authors of \cite{ZWCr07} described the number of cuspidal configurations for discrete values of $\r_1$ in $\mathbb{R}$. This allowed them to find configurations that were omitted in previous works.

In this section, we give a certified and exhaustive description of the number of cuspidal configurations as function of $\r_1$. In particular, this study allows us to find values of $\r_1$, for which the 3-R\underline{P}R manipulator has $10$ cuspidal configurations, while previous works never observed more than $8$ cuspidal configurations. This gives new possibilities to change assembly mode without crossing singularities. More generally, our method compute the number of cuspidal configurations for all the values of $\r_1$ in open intervals, and not only slices.

The dimension of the set of solutions of Equations \ref{fullcuspidal} in the complex field is $1$. This can be verified by computing and analyzing the Gr\"obner bases of Equations  \ref{fullcuspidal} (see \cite[chapter 9]{CLO92} for more details).

To describe geometrically the solutions of Equations \ref{fullcuspidal} as function of $\r_1$, we consider it as a parametric system where:
\begin{itemize}
  \item the single parameter is $\r_1$
  \item the unknowns are $\r_2,\r_3,\U_x,\U_y,\t_x,\t_y$
\end{itemize}

First, using \cite[page[341]{BWbook93}, we can check that for almost all values of $\r_1$, the roots of Equations \ref{fullcuspidal} have multiplicity $1$ and are thus corresponding exacty to cuspidal configurations.

Then we follow the work of \cite{LR07, CR02, FMRS08} to describe the roots of a parametric system. Our process has $2$ steps:
\begin{itemize}
  \item we first compute its \emph{minimal discriminant variety}.
  \item we then compute the number of solutions for sample parameter's values chosen outside the discriminant variety
\end{itemize}
\subsubsection{Discriminant variety of the cuspidal configurations}
We consider Equations \ref{fullcuspidal} as a system parametrized with $\r_1$. In this case, its minimal discriminant variety is a finite set of values of $\r_1$ denoted by:
$$
\DV=\{a_1,\dots,a_k\}\subset\mathbb{R}, k>0
$$

The main property of the discriminant variety is that for each value of $\r_1$ in an interval $]a_i,a_{i+1}[$, the Equations \ref{fullcuspidal} has the same number of distinct real solutions.

The points of the discriminant varieties are the real roots of univariate polynomials. The lines $\r_1$ of table \ref{numdv} show numerical approximations of these roots. The full univariate polynomials are not given here for lack of space. They can be found at the address \url{http://www.irccyn.ec-nantes.fr/~chablat/3RPR_cuspidal.html}. They were computed in 11 seconds with a 2.9GHz Intel cpu.

\begin{table}
  \begin{center}
    \begin{tabular}{|c|c@{}c@{}c@{}c@{}c@{}c@{}c@{}c|}
  & & & & & & & & \\
\hline \#Cusp&&0&&2&&4&&2\\
      $\r_1$&
      0.000 &]------[&
      0.148&]------[&
      1.655&]------[&
      1.660&]------[\\[1em]
      \hline \#Cusp&&4&&6&&8&&6\\
      $\r_1$&
      2.261&]------[&
      2.975&]------[&
      9.186&]------[&
      9.186$^*$&]------[\\[1em]
      \hline \#Cusp&&8&&6&&8&&6\\
      $\r_1$&
      9.257&]------[&
      9.257$^*$&]------[&
      10.905&]------[&
      10.905$^*$&]------[\\[1em]
      \hline \#Cusp&&8&&6&&8&&6\\
      $\r_1$&
      14.579&]------[&
      14.579$^*$&]------[&
      20.555&]------[&
      20.562&]------[\\[1em]
      \hline \#Cusp&&8&&10&&8&&6\\
      $\r_1$&
      26.786&]------[&
      28.094&]------[&
      28.107&]------[&
      28.257&]------[\\[1em]
      \hline \#Cusp&&8&&6&&\multicolumn{3}{c|}{4}\\
      $\r_1$&
      30.740&]------[&
      30.779&]------[&
      30.946&\multicolumn{3}{@{}l|}{]------------------------}
      \\[1em]
      \hline
    \end{tabular}\\[\baselineskip]
  \end{center}
    $^*$ these values are close but distinct, the closest values differ by at least $10^{-11}$.
    \caption{Discriminant variety of the cuspidal configurations (Equations  \ref{fullcuspidal}) w.r.t. $\r_1$; the numerical values were truncated}
    \label{numdv}
\end{table}

The lines \#Cusp gives the number of cuspidal configurations for $\r_1$ in an interval $]a_i,a_{i+1}[$, where the $a_i$ are the real values of $\r_1$ defining the discriminant variety $\DV$.
\subsubsection{Number of cuspidal configurations}
Using the main property of the discriminant variety, we know that to count the number of cuspidal configurations outside the discriminant variety, it is sufficient to
compute a finite set of sampling points in the complement of $\DV$ and to solve the corresponding zero-dimensional systems. More precisely, we choose a value $v_i$ in each interval:
$$
]a_i,a_{i+1}[, \mbox{ such that $a_i,a_{i+1}$ are consecutive reals of \DV }
$$
and compute the number of solutions to Equations \ref{fullcuspidal} when $\r_1=v_i$. Moreover, we choose a value $v_{k+1}$ in $]a_k,+\infty[$, and count the number of solutions of Equations \ref{fullcuspidal} when $\r_1=v_{k+1}$. The results of these computations are summarized in the lines \#Cusp of Table \ref{numdv}. To count the number of real solutions of Equations \ref{fullcuspidal} when $\r_1=v_i$, we use the real solver given by \rsisolate in Maple. It took around 5 minutes to solve the systems induced by each of the 23 sample points with a 2.9GHz Intel cpu. 

\begin{figure}[ht]
\begin{center}
    \includegraphics[scale=1]{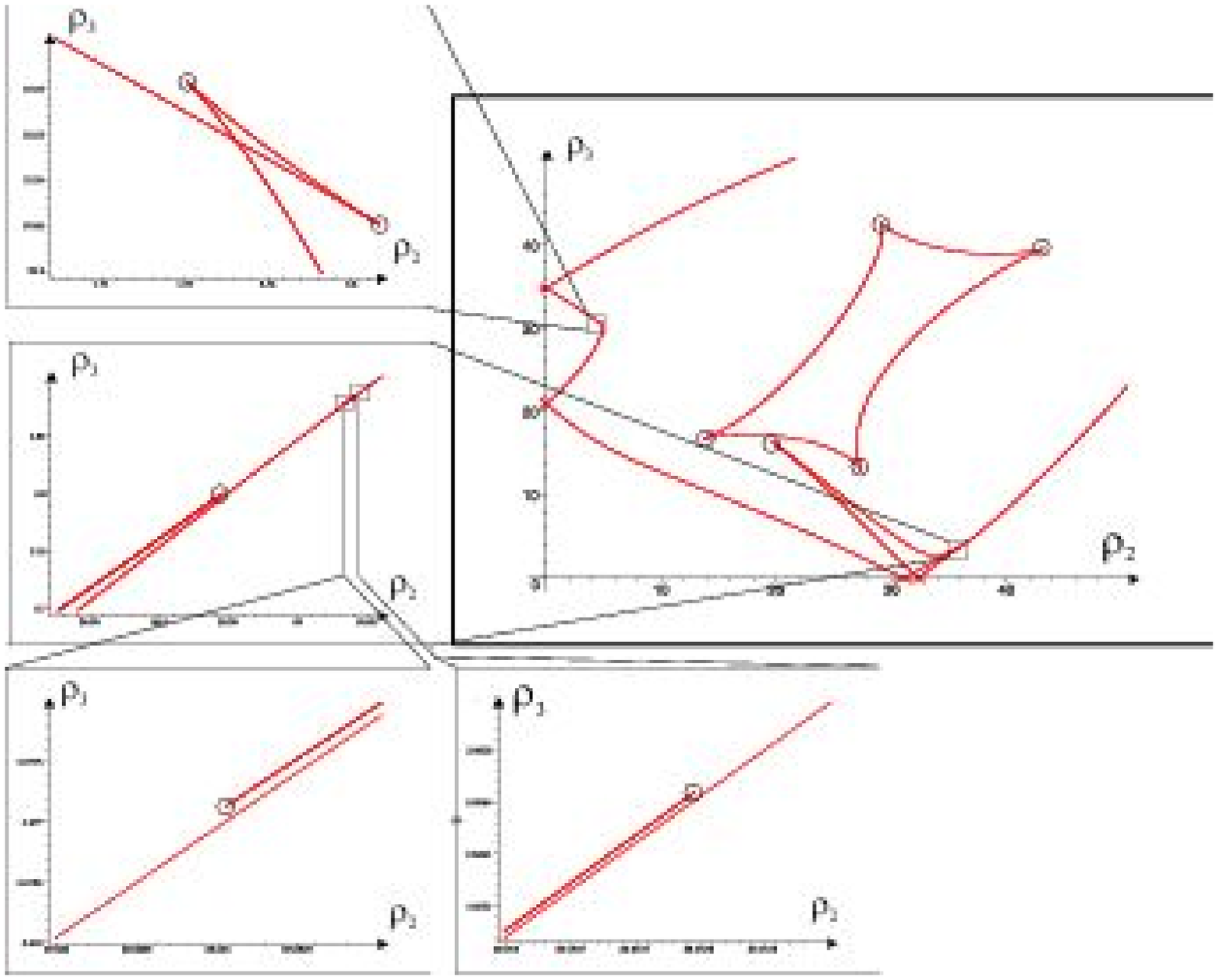}
\caption{Singular curve for $\r_1=28.10$. The circles show the 10 cuspidal configurations in this configuration.}
\label{2810}
\end{center}
\end{figure}

From Table \ref{numdv} we conclude that, asymptotically, the manipulator has $4$ cuspidal configurations, and that this number is stable as soon as $\r_1$ is greater than $31$. Moreover, we can observe that the robot may have up to 10 cuspidal configurations when $\r_1$ is in the interval $]28.095,28.107[$. Figure \ref{2810} shows the singular curve of the parallel robot for $\r_1=28.10$ and the corresponding 10 cuspidal configurations.

\section{Conclusion}
This paper shows that efficient algebraic tools can be applied to analyse in a certified way important kinematic features of parallel manipulators such as the determination of cusp points. These points are known to play an important role in planning non-singular assembly mode changing motions.

A new method was introduced, which is able to characterize all the cusp points for the 3-R\underline{P}R manipulators. It allowed us to determine the number of cusp points for all the slices of the joint space. For the first time, a section for a given $\rho_1$ was found with 10 cusp points.

Table \ref{algorithms} summarizes the method used in this article and relates it to the 2 previous main methods to compute cuspidal configurations in the literature.

This results can be computed with a toolbox available now in Maple 12 and later versions \cite{LGJMcca09}.

\begin{table}

\begin{tabular}{|c||c|c||c|}
\hline
\bf
\begin{tabular}{c}Original\\System\end{tabular}

& \multicolumn{3}{c@{}|}{
  $\gbf{E} \left\{\begin{array}{c}
     p_1(\gbf{L},x_1,...,x_n)=0\\
     \vdots\\
     p_n(\gbf{L},x_1,...,x_n)=0\\
     \end{array}\right.$
\hfill\small
  $      \begin{array}{|l@{}l}
            \mbox{MD Model: }& \left\{\begin{array}{l@{}l}
                 \mbox{parameters: }& \gbf{L}\\
                 \mbox{unknowns: }& x_1,...,x_3\\[-0.5em]
                 &\hfill:= \theta_1, \theta_2,\theta_3
                 \end{array}\right.
                 \\
                 \hline
            \mbox{Model GRS: }& \left\{\begin{array}{l@{}l}
                 \mbox{parameters: }& \gbf{L}\\
                 \mbox{unknowns: } & x_1,...,x_4\\[-0.5em]
                 &\hfill:= {B_1}_x, {B_1}_y,\alpha_x,\alpha_y\\
                 \end{array}\right.
                 \\
            \end{array}$
}
\\
\hline

& \begin{tabular}{c}\textbf{Elimination}\\
                    (\cite{OW95,CR02,Wjmd98})\end{tabular}

& \begin{tabular}{c}\textbf{Series Expansion}\\
                    (\cite{McaD99,ZWCr07})\end{tabular}

& \begin{tabular}{c}\textbf{Jacobian}\\
                    (this article)\end{tabular}

\\[2em]
\cdashline{2-4}
\bf
\begin{tabular}[b]{c}
Method\\[1em]
\\
\end{tabular}

& $p_0(\rho_1,...,\rho_s,x_1)$

&
$
\begin{array}{l}
\Delta \gbf{E}_i = \\[1em]
\left(
\sum_{j=1}^{n} \Delta \theta_i \frac{\partial}{\partial \theta_j}\right.\\ +
\left.\sum_{j=1}^{n} \Delta L_i \frac{\partial}{\partial \rho_j}
\right) \gbf{E}_i\\[1em]
+ \frac{1}{2!}
\left(
\sum_{j=1}^{3} \Delta \theta_i \frac{\partial}{\partial \theta_j}\right.\\ +
\left.\sum_{j=1}^{3} \Delta L_i \frac{\partial}{\partial \rho_j}
\right)^2 \gbf{E}_i\\
\end{array}$

& $J := \left(\begin{array}{c}
  \frac{\partial p_1}{\partial x_1} \cdots \frac{\partial p_1}{\partial x_n}\\
  \vdots\\
  \frac{\partial p_n}{\partial x_1} \cdots \frac{\partial p_n}{\partial x_n}\\
\end{array}\right)$
\\
\hline
\bf
\begin{tabular}{c}Singular\\Config.\end{tabular}

& $(S) \left\{\begin{array}{l}\gbf{E}=0\\
          \frac{\partial p_0}{\partial x_1}=0\end{array}\right.$

& $(S) \left\{\begin{array}{l}
    \gbf{E}=0\\
    \mbox{first order}\\
    \mbox{ matrix drops rank}
  \end{array}\right.$

& $(S) \left\{\begin{array}{l}\gbf{E}=0\\\J=0\end{array}\right.\hfill$

\\
\hline
\bf
\begin{tabular}{c}Cuspidal\\Config.\end{tabular}

& $(C) \left\{\begin{array}{l}
     \gbf{E}=0  \\
     \frac{\partial p_0}{\partial x_1}=0 \\
     \frac{\partial^2 p_0}{\partial x_1^2}=0 \\
  \end{array}\right.$

& $(C)  \left\{\begin{array}{l}
    \gbf{E}=0 \\
    \mbox{first order}\\
    \mbox{second order}\\
    \mbox{ matrices drop rank}
  \end{array}\right. $

& $(C)  \left\{\begin{array}{l}
    \gbf{E}=0\\
    \J =0\\
    \left(\begin{array}{c}
  \frac{\partial J}{\partial x_1} \cdots  \frac{\partial J}{\partial x_n}\\
  \frac{\partial p_1}{\partial x_1} \cdots  \frac{\partial p_1}{\partial x_n}\\
  \vdots\\
  \frac{\partial p_n}{\partial x_1} \cdots  \frac{\partial p_n}{\partial x_n}\\
\end{array}\right)\\
    \hfill\mbox{drops rank}
  \end{array}\right.\hfill$

\\

\hline
\bf
\begin{tabular}{c}
Description\\of Cuspidal\\Config.
\end{tabular}
&
\multicolumn{3}{c|}{ Discriminant Variety of $(C)$}\\

\hline

\end{tabular}

\caption{The two main methods and one original methods for singular and cuspidal configurations characterization. Only the jacobian method allowed us to describe completely the cuspidal configurations.}
\label{algorithms}
\end{table}

\section*{Acknowledgement}
We would like to thank the reviewers for their useful remarks, which we have all taken into account.

\bibliographystyle{unsrt}
\bibliography{citations}
\end{document}